\documentclass{article}

\usepackage{microtype}
\usepackage{graphicx}
\usepackage{subcaption}
\usepackage{booktabs} 

\usepackage{hyperref}

\usepackage[accepted]{icml2026}

\usepackage{amsmath}
\usepackage{amssymb}
\usepackage{mathtools}
\usepackage{amsthm}
\usepackage{listings}

\usepackage{xcolor}
\definecolor{keywordcolor}{rgb}{0.7, 0.1, 0.1}   
\definecolor{tacticcolor}{rgb}{0.0, 0.1, 0.6}    
\definecolor{commentcolor}{rgb}{0.3, 0.5, 0.3}   
\definecolor{symbolcolor}{rgb}{0.0, 0.1, 0.6}    
\definecolor{sortcolor}{rgb}{0.1, 0.5, 0.1}      
\definecolor{rulecolor}{rgb}{0, 0, 0}
\definecolor{attributecolor}{rgb}{0.7, 0.1, 0.1} 

\usepackage[capitalize,noabbrev]{cleveref}

\theoremstyle{plain}

\theoremstyle{definition}

\theoremstyle{remark}

\usepackage[textsize=tiny]{todonotes}

\begin{document}

\onecolumn
\icmltitle{Numina-Lean-Agent: An Open and General Agentic Reasoning System\\for Formal Mathematics}



\icmlsetsymbol{equal}{*}

\begin{icmlauthorlist}
\icmlauthor{Junqi Liu}{equal,amss}
\icmlauthor{Zihao Zhou}{equal,UoL,xjtlu}
\icmlauthor{Zekai Zhu}{equal,tongji}
\icmlauthor{Marco Dos Santos}{cam}
\icmlauthor{Weikun He}{amss}
\icmlauthor{Jiawei Liu}{numina}
\icmlauthor{Ran Wang}{numina}
\icmlauthor{Yunzhou Xie}{IC}
\icmlauthor{Junqiao Zhao}{tongji}
\icmlauthor{Qiufeng Wang}{xjtlu}
\icmlauthor{Lihong Zhi}{amss}
\icmlauthor{Jia Li}{numina}
\icmlauthor{Wenda Li}{Edinburgh}
\end{icmlauthorlist}

\icmlaffiliation{amss}{Academy of Mathematics and Systems Science, University of Chinese Academy of Sciences\\}
\icmlaffiliation{tongji}{Tongji University\\}
\icmlaffiliation{cam}{University of Cambridge~~}
\icmlaffiliation{IC}{Imperial College London~~}
\icmlaffiliation{Edinburgh}{University of Edinburgh}
\icmlaffiliation{UoL}{University of Liverpool~~}
\icmlaffiliation{xjtlu}{Xi'an Jiaotong-Liverpool University~~}
\icmlaffiliation{numina}{Project Numina~~}

\icmlcorrespondingauthor{Jia Li}{jia@projectnumina.ai}
\icmlcorrespondingauthor{Wenda Li}{wenda.li@ed.ac.uk}


\vskip 0.3in

\printAffiliationsAndNotice{\icmlEqualContribution}

\begin{abstract}
Agentic systems have recently become the dominant paradigm for formal theorem proving, achieving strong performance by coordinating multiple models and tools. However, existing approaches often rely on task-specific pipelines and trained formal provers, limiting their flexibility and reproducibility.
In this paper, we propose the paradigm that directly uses a general coding agent as a formal math reasoner. This paradigm is motivated by (1) A general coding agent provides a natural interface for diverse reasoning tasks beyond proving, (2) Performance can be improved by simply replacing the underlying base model, without training., and (3) MCP enables flexible extension and autonomous calling of specialized tools, avoiding complex design.  Based on this paradigm, we introduce \textbf{Numina-Lean-Agent}, which combines Claude Code with Numina-Lean-MCP to enable autonomous interaction with Lean, retrieval of relevant theorems, informal proving and auxiliary reasoning tools.
Using Claude Opus 4.5 as the base model, Numina-Lean-Agent solves all problems in Putnam 2025 (12/12), matching the best closed-source system. Beyond benchmark evaluation, we further demonstrate its generality by interacting with mathematicians to successfully formalize the Brascamp–Lieb theorem. We release Numina-Lean-Agent and all solutions at \href{https://github.com/project-numina/numina-lean-agent}{https://github.com/project-numina/numina-lean-agent}.
\end{abstract}

\section{Introduction}
Formal theorem proving aims to construct machine-verifiable proofs for mathematical theorems within rigorously defined logical systems, such as Lean~\yrcite{de2015lean} and Isabelle~\citep{paulson1994isabelle}. Unlike informal mathematical reasoning, formal verification systems provide tools for automatically and soundly verifying the correctness of proofs. Consequently, these systems establish a foundation for developing reliable reasoning.
Previous advances in neural theorem proving have focused on developing single-model formal provers. Early provers relied on tactic prediction combined with explicit search methods, such as Monte Carlo tree search, to explore the proof space.~\citep{lample2022hypertree, alphaproof}. To mitigate the efficiency limitations of search-based methods, subsequent work explored whole proof generation to directly produce complete proofs~\citep{xin2024deepseekproverv15harnessingproofassistant}. Subsequently, other efforts incorporated informal reasoning to guide tactic generation and proof construction~\citep{wang2025kiminaproverpreviewlargeformal, ren2025deepseek}. Despite notable progress, effectively capturing long-horizon, structured reasoning within formal systems remains a central challenge.

More recently, several systems have moved beyond single-model formal provers by introducing agentic workflows that enable provers to interact with formal theorem proving environments and other models. For example, HILBERT~\citep{varambally2025hilbert} proposes an agentic framework that combines informal reasoners with formal provers to guide proof construction. In parallel, Seed-Prover 1.5~\citep{chen2025seed15} trains a formal prover via large-scale agentic reinforcement learning, emphasizing repeated interaction with the Lean compiler and related tools. In addition, AxiomProver~\citep{axiommath2025seeingwhy}, developed by Axiom Math, adopts an autonomous multi-agent ensemble architecture and has achieved a perfect score on Putnam-2025. These systems highlight the growing effectiveness of agentic proving systems.
Despite their strong performance, existing agentic proving systems exhibit several limitations: (1) They rely on task-specific reasoning pipelines that are explicitly designed and often coupled with extensively trained formal provers, which can limit their extensibility to new tools or domains. (2) Most systems are closed-source and provide limited implementation details, making it difficult for the broader community to reproduce and extend their work.

In this paper, we propose a paradigm for building a formal math reasoner based on a general coding agent, which is motivated by three reasons: (1) Coding agents provide a native interface for diverse proof engineering tasks beyond proving. (2) It allows for the flexible replacement of underlying base models to enhance reasoning capabilities without any training. (3). Integration of MCP enables the plug-and-play extension of specialized reasoning tools and the model can autonomously invoke them based on the specific query.
Following this paradigm, we propose Numina-Lean-Agent, which consists of Claude Code and Numina-Lean-MCP for providing diverse reasoning tools. 
Specifically, Numina-Lean-MCP integrates several core components.
It includes \textit{Lean-LSP-MCP}~\citep{lean-lsp-mcp}~\footnote{\url{https://github.com/oOo0oOo/lean-lsp-mcp}}
 for agentic interaction with the Lean theorem prover.
To provide relevant background knowledge, we develop \textit{LeanDex} for semantic retrieval of related theorems and definitions from Lean libraries such as mathlib. An \textit{Informal Prover}~\citep{huang2025winning} is used to generate detailed informal proof solutions, while a \textit{Discussion Partner} tool enables querying external language models to assist in reasoning and planning.
Together, these components make Numina-Lean-Agent a general and powerful formal mathematical reasoning system.

Using Claude Opus 4.5~\citep{ClaudeOpus45} as the base model, Numina-Lean-Agent successfully solved all 12 problems in the Putnam 2025, achieving state-of-the-art performance. This result matches the closed-source system AxiomProver and surpasses Harmonic’s Aristotle~\citep{aristotle} by two problems. We report all solutions along with their computational costs and proof lengths in Section~\ref{sec3}. Notably, on certain problems such as Problem-B1, Numina-Lean-Agent produced remarkably concise proofs than AxiomProver and Seed-Prover 1.5. 
Beyond standard automated proving, Numina-Lean-Agent serves as a general mathematical reasoning system, enabling mathematicians to engage in interactive "vibe proving". We demonstrate this paradigm by collaborating with human experts to formalize the Brascamp-Lieb theorem, with the details of the interactive process reported in Section~\ref{sec4}.

\begin{figure*}[t]
    \centering
    \includegraphics[scale=0.47]{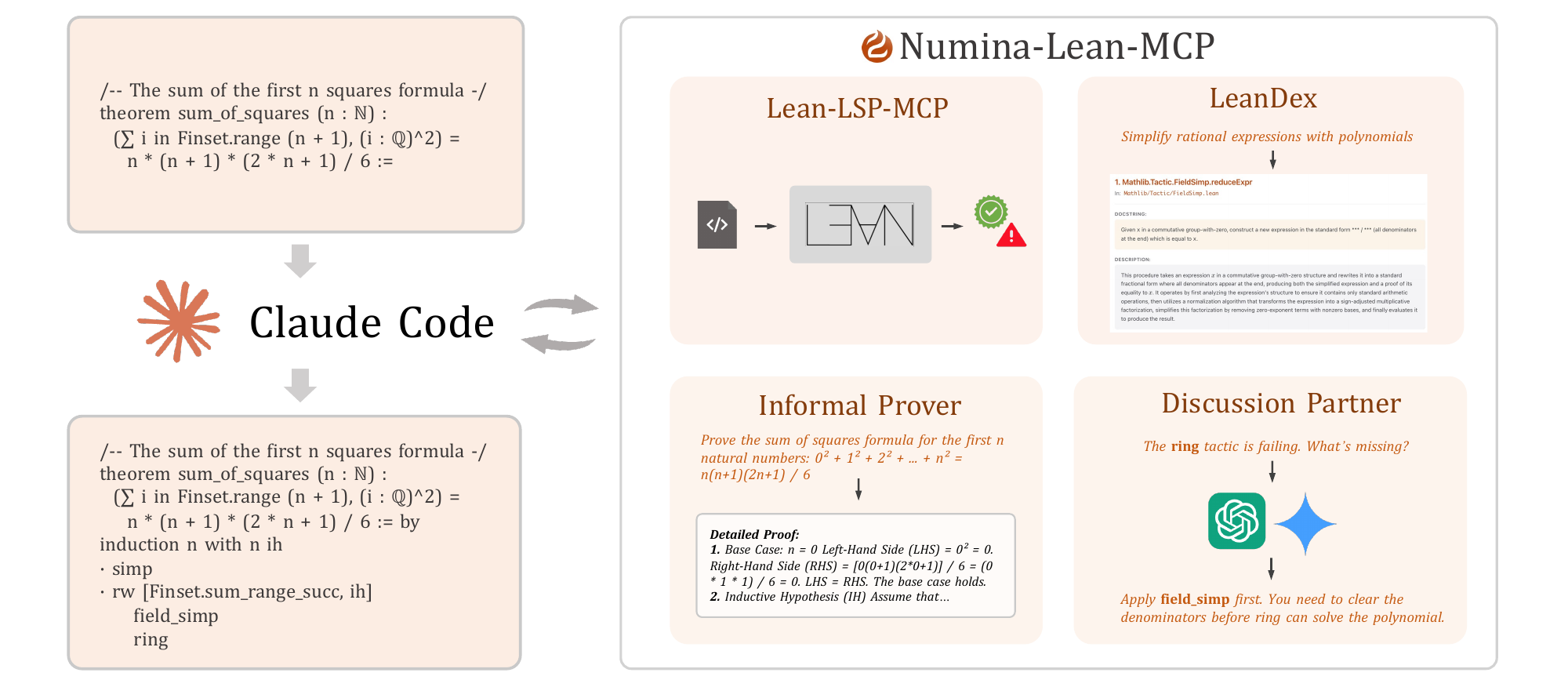}
    \caption{Overview of Numina-Lean-Agent, an agentic formal reasoning framework built on Claude Code and Numina-Lean-MCP. The agent autonomously selects and invokes specialized reasoning tools to handle diverse queries. }
    \label{fig:overview}
\end{figure*}

\section{Numina-Lean-Agent}
\subsection{Overview}

As shown in Figure~\ref{fig:overview}, Numina-Lean-Agent is an agentic formal theorem proving framework built upon Claude Code and Numina-Lean-MCP. Functioning as an autonomous agent, it can dynamically select and invoke the appropriate reasoning tools within Numina-Lean-MCP to handle diverse queries and complete complex formal reasoning tasks.

\subsection{Numina-Lean-MCP}
\textbf{Lean-LSP-MCP. } Lean-LSP-MCP~\citep{lean-lsp-mcp} is a Model Context Protocol (MCP) server explicitly designed for the Lean theorem prover. Acting as a bridge between LLMs and the Lean kernel via the Language Server Protocol (LSP), it empowers models with the capability to deeply comprehend, analyze, and manipulate Lean projects. It significantly enhances the proving capabilities of models through a toolset organized into three distinct dimensions:

\begin{enumerate}
    \item Semantic Awareness and Interaction: This dimension provides a suite of tools enabling agents to simulate the behavior of proficient Lean users. Ranging from \texttt{lean\_file\_outline} for grasping global structures, to \texttt{lean\_goal} for precise goal querying, and \texttt{lean\_diagnostic\_messages} for obtaining authoritative verification results, these tools liberate models from guessing regarding proof states, enabling precise decision-making grounded in the authentic compilation environment.
    
    \item Code Execution and Strategy Exploration: It supports the instant compilation of isolated code snippets via \texttt{lean\_run\_code} and utilizes \texttt{lean\_multi\_attempt} to allow parallel execution and evaluation of multiple strategies at a single proof node. This mechanism establishes a robust ``trial-feedback-optimization'' closed-loop for automated theorem proving.
    
    \item Theorem Retrieval and Knowledge Enhancement: Multi-level search tools are integrated to effectively mitigate hallucinations. \texttt{lean\_local\_search} focuses on mining definitions within local lean projects and the standard library (stdlib), while \texttt{lean\_loogle} facilitates searching the massive Mathlib repository via natural language or structured queries. This dual-retrieval mechanism ensures that every theorem cited by the model is factually existent and contextually appropriate.
\end{enumerate}

\textbf{LeanDex.} We present a new theorem search tool for Lean that supports theorem retrieval under Lean v4.26.0. In contrast to existing tools, loogle imposes strict requirements on the format of search queries, while local search is mainly limited to searching within a local project. It is an agentic semantic search tool for Lean, capable of retrieving mathematical theorems and definitions across multiple packages, including mathlib and FLT. Given a natural language query, Leandex employs an intelligent agent to interpret and reason about the query, identifying the most relevant Lean objects. Built on top of LeanExplore, it extends the underlying semantic search framework with enhanced reasoning and retrieval capabilities, significantly improving both flexibility and coverage.

\textbf{Informal Prover. }
We implement a lightweight Gemini IMO agent system~\citep{huang2025winning} as the Informal Prover to generate detailed informal solutions. The system consists of two models: a Generator and a Verifier. The Generator is responsible for generating informal solutions, while the Verifier assesses the correctness of the generated solutions. These two models interact in an iterative refinement loop. When the Verifier identifies errors in the generated proof, it will provide feedback to the Generator. In the next iteration, the Generator refines its solution base on both the previous solution and the Verifier's feedback. This process continues until the Verifier accepts the solution as correct or a maximum number of iterations is reached, which we set to 20.

To improve the reliability of verification, the Verifier evaluates each candidate solution independently three times. A solution is accepted only if all three verification passes judge it to be correct. In our implementation, we use \textit{Gemini-3-Pro-Preview} for both the Generator and the Verifier.

\textbf{Discussion Partner.} In scientific research, discussion is widely recognized as a highly effective cognitive tool. By exchanging diverse viewpoints and reasoning paths, researchers often overcome mental blind spots and spark new inspiration, thereby facilitating problem-solving. Inspired by this insight, we designed and implemented \texttt{discuss\_partner}, a tool designed to assist the automated formalization process.

Specifically, this tool empowers Claude Code with the ability to 'seek assistance' during Lean formalization: when encountering obstacles—such as proof bottlenecks, dilemmas in strategy selection, or ambiguities in intermediate lemmas—the primary model can proactively initiate discussions with other LLMs. These collaborating models analyze the current state from distinct reasoning perspectives, offering candidate ideas or alternative proof paths to provide insightful feedback. Leveraging this multi-model collaborative mechanism, the system can explore the proof space more effectively, significantly enhancing both the robustness and the success rate of the formalization process.

\section{Evaluation}
\label{sec3}

\subsection{Performance}

We evaluated Numina-Lean-Agent on the Putnam 2025 benchmark and compared its performance with other existing provers. Notably, we used the formal statements provided by Seed-Prover 1.5. Moreover, all operations executed by our agent were strictly sequential, without any form of parallelization. Internet search access was disabled for all API calls for preventing the agent from retrieving solutions via online search. Under these settings, Numina-Lean-Agent achieved state-of-the-art performance, successfully solving 12 out of 12 problems on Putnam 2025. By default, each problem is allocated an approximate budget of \$50. Due to their substantially higher difficulty and longer proof search trajectories, problem A5 is assigned a larger budget of approximately \$1000, and problem B6 is allocated an increased budget of approximately \$300. These values are intended to reflect relative computational effort rather than exact accounting.

\begin{table}[tbh]
    \centering
    \caption{Performance comparison of Numina-Lean-Agent against other methods}
    \label{tab:result_comp}
        \begin{sc}
        \begin{tabular}{lcccccccccccc}
            \toprule
            \textbf{Putnam2025} & A1 & A2 & A3 & A4 & A5 & A6 & B1 & B2 & B3 & B4 & B5 & B6 \\
            \midrule
            Aristotle & \checkmark & \checkmark & \checkmark & \checkmark & &\checkmark & \checkmark& \checkmark& \checkmark & & \checkmark& \checkmark \\
            Seed-Prover 1.5 & \checkmark & \checkmark & \checkmark & \checkmark & &\checkmark & \checkmark& \checkmark& \checkmark & \checkmark & \checkmark & \checkmark \\
            Axiom & \checkmark & \checkmark & \checkmark & \checkmark &
            \checkmark & \checkmark & \checkmark & \checkmark & \checkmark & \checkmark & \checkmark & \checkmark \\
            \midrule
            Numina-Lean-Agent & \checkmark & \checkmark & \checkmark & \checkmark &
            \checkmark & \checkmark & \checkmark & \checkmark & \checkmark & \checkmark & \checkmark & \checkmark \\
            \bottomrule 
        \end{tabular}
        \end{sc}
\end{table}

We conducted a comparative experiment on two design paradigms for the \verb|informal_prover| tool using the Putnam 2025 B4 problem. The first approach employs an iterative refinement strategy: generating an initial solution, verifying it, revising based on feedback, and repeating this cycle for n iterations~\citep{huang2025winning}. The second approach adopts an independent sampling strategy: generating n solutions independently and verifying each separately. The total number of calls remains identical across both approaches to ensure a fair comparison. Experimental results demonstrate that, under the same search configuration, the iterative refinement strategy significantly outperforms independent sampling. Specifically, independent sampling failed to complete the formal proof of B4 within 10 rounds, whereas iterative refinement successfully completed the proof in only 5 rounds, clearly demonstrating the advantage of feedback-driven iterative correction in improving proof efficiency. We also propose a new subagent-based approach to solve problem A5; the detailed methodology will be discussed in~\cref{putnam_a5}.

\begin{table}[tbh]
    \centering
    \caption{Evaluation results on Putnam 2025 by Numina-Lean-Agent.}
    \label{tab:main_results}
        \begin{sc}
        \begin{tabular}{lcccccccccccc}
            \toprule
            \textbf{Putnam2025} & A1 & A2 & A3 & A4 & A5 & A6 & B1 & B2 & B3 & B4 & B5 & B6 \\
            \midrule
            w/o informal & \checkmark & \checkmark & & \checkmark & & & & & \checkmark & & & \\
            w informal & \checkmark & \checkmark & \checkmark & \checkmark &
            & \checkmark & \checkmark & \checkmark & \checkmark & \checkmark & \checkmark & \checkmark \\
            w subagent & \checkmark & \checkmark & \checkmark & \checkmark &
            \checkmark & \checkmark & \checkmark & \checkmark & \checkmark & \checkmark & \checkmark & \checkmark \\
            \bottomrule \\
        \end{tabular}
        \end{sc}
\end{table}

As shown in~\cref{tab:time_comp}, we compare the per-problem solving time of Numina-Lean-Agent against other representative provers on Putnam 2025. Despite the fact that Numina-Lean-Agent operates without any parallel execution, it demonstrates notable efficiency advantages on a subset of problems, achieving shorter solving times than competing methods on several instances.

\begin{table}[tbh]
    \centering
    \caption{Time spent comparison of Numina-Lean-Agent against other methods (Unit: minutes)}
    \label{tab:time_comp}
        \begin{sc}
        \begin{tabular}{lcccccccccccc}
            \toprule
            \textbf{Putnam2025} & A1 & A2 & A3 & A4 & A5 & A6 & B1 & B2 & B3 & B4 & B5 & B6 \\
            \midrule
            Aristotle & {\bf 30} & 60 & {\bf 30} & 180 & -- & {\bf 60} & 150 & {\bf 25} & 40 & -- & 420 & {\bf 180} \\
            Seed-Prover 1.5  & 60 & {\bf 30} & 120 & 240 & -- & 240 & 540 & 360 & {\bf 30} & 120 & 240 & {\bf 180} \\
            Axiom & 110 & 180 & 165 & {\bf 107} & {\bf 518}  & 259 & 270 & 65 & 43 & {\bf 112} & 254 & 494 \\
            \midrule
            Numina-Lean-Agent & 97 & {\bf 30} & 44 & 169 & 2040 & 89 & {\bf 55} & 142 & {\bf 30} & 308 & {\bf 88} & 797 \\
            \bottomrule
        \end{tabular}
        \end{sc}
\end{table}

In~\cref{tab:code_length}, we further compare the proof length generated by different provers. For a fair comparison, we remove all comments and blank lines from the final Lean code and measure the resulting number of lines. The results show that, compared with AxiomProver and Seed-Prover~1.5, Numina-Lean-Agent produces shorter proofs on a substantial number of problems; in particular, on A3, B1, and B5, the advantage is especially pronounced. We note that step-based provers have an inherent advantage in producing very short proofs, and therefore the proofs generated by Numina-Lean-Agent are generally longer than those produced by Aristotle. Nevertheless, when compared with other agentic provers under a similar setting, Numina-Lean-Agent consistently yields more concise formalizations on most problems, demonstrating its effectiveness in generating compact and efficient formal proofs.

\begin{table}[tbh]
    \centering
    \caption{Code length comparison of Numina-Lean-Agent against other methods}
    \label{tab:code_length}
        \begin{sc}
        \begin{tabular}{lcccccccccccc}
            \toprule
            \textbf{Putnam2025} & A1 & A2 & A3 & A4 & A5 & A6 & B1 & B2 & B3 & B4 & B5 & B6 \\
            \midrule
            aristotle & 45 & 195 & 103 & 291 & -- & 123 & 223 & 108 & 70 & -- & 291 & 280 \\
            Seed-Prover 1.5 & 631 & 469 & 927 & 1095 & -- & 881 & 849 & 1613 & 584 & {\bf 628} & 2499 & 2594 \\
            Axiom & 556 & 458 & 1089 & 825 & {\bf 1878} & {\bf 468} & 1179 & {\bf 346} & 302 & 993 & 1310 & {\bf 862} \\
            \midrule
            Numina-Lean-Agent & {\bf 365} & {\bf 401} & {\bf 422} & {\bf 605} & 3263 & 835 & {\bf 328} & 690 & {\bf 292} & 648 & {\bf 929} & 1820 \\
            \bottomrule
        \end{tabular}
        \end{sc}
\end{table}

\subsection{Putnam-2025-A5}\label{putnam_a5}

For problem A5, we adopt a novel subagent mechanism that decomposes the proof into several subgoals and solves them independently, effectively mitigating the issue of excessively long contexts. Our empirical observations indicate that when the context becomes too long, the model’s ability to follow instructions degrades significantly, making it difficult to focus on a single proof objective, which in turn hampers the resolution of critical subgoals. By introducing subagents and modularizing the proof task, we can substantially alleviate these issues and improve the overall proof efficiency.

The core of A5 is to prove that, among all permutations satisfying a certain property, alternating permutations occur in the largest number. In several previous experiments, the model repeatedly got stuck on this key lemma. We conjecture that this difficulty is caused by overly long contexts, and therefore adopt a subagent strategy that isolates this lemma from the overall proof and handles it separately. Concretely, the subagent first invokes GPT-5.2 to generate an informal hint, and then carries out the corresponding formalization guided by this hint. This process can be iterated until the lemma is successfully formalized.

\section{Formalizing Brascamp Lieb with Numina-Lean-Agent}
\label{sec4}

\subsection{Blueprint Generation}
\label{sec:blueprint}

Formalizing a complex theorem in Lean is a long-horizon task with dense dependencies. When Claude Code is asked to directly prove the final statement, it often commits to suboptimal formulations and gets trapped in local dead ends. We therefore introduce a \textbf{blueprint} as an explicit planning layer that decomposes the global goal into a sequence of verifiable subgoals.

A blueprint is a design-document-style artifact consisting of (i) required definitions and notation, (ii) a curated list of intermediate lemmas with suitable granularity, and (iii) the final theorem whose proof largely composes these lemmas. Dependencies are recorded explicitly (e.g., via \verb|\uses{...}|), forming a DAG that determines proving order and reduces ambiguity during search.

Importantly, blueprint generation is \emph{recursive} and tightly coupled to the formalization loop rather than a one-shot preprocessing step. As the agent attempts to discharge lemmas in Lean, compilation feedback and proof-state inspection may reveal that an informal step is incorrect, underspecified, or split at an unsuitable granularity. In such cases, the agent revisits and refines the blueprint (e.g., strengthening assumptions, rephrasing statements to match Lean interfaces, or inserting missing intermediate lemmas) and then continues formalization with the updated plan. To improve robustness, the agent can also invoke external discussion models (e.g., Gemini) to propose alternative decompositions or repairs when the current blueprint repeatedly leads to bottlenecks.

Overall, the blueprint plays the role of a high-level mathematical plan: a stronger mathematical reasoner is used to decompose a difficult statement into a sequence of small, checkable steps, while Claude Code focuses on turning these steps into machine-verifiable Lean proofs. Crucially, verification is not merely an endpoint—Lean feedback (failed typeclass search, missing lemmas, mismatched interfaces, etc.) provides concrete signals that are fed back to revise the blueprint, yielding a closed-loop ``plan--formalize--refine'' workflow that stabilizes long-horizon formalization.

\subsection{Human-AI Cooperation}

We design a human–machine collaborative interaction framework for Numina-Lean-Agent, enabling human experts to work together with the agent by writing hints and modifying the Blueprint. One of the authors of this paper is a mathematician, and we conduct a collaborative case study based on his preprint~\textbf{Effective Brascamp–Lieb Inequalities}~\yrcite{benard2025effective}, published in November 2025. We show the formal statement of the main theorem of the Effective Brascamp–Lieb inequalities in Appendix~\ref{fs_of_BL}. In this experiment, a mathematician, a Lean formalization expert, and Numina-Lean-Agent jointly cooperate to formalize the results of the paper.

Over a period of less than two weeks of intermittent collaboration, the two human experts and the agent completed the formalization of more than \textbf{8,000} lines of Lean code. During this process, the agent autonomously introduced approximately \textbf{70} new definitions, lemmas, and theorems, illustrating its ability to actively extend the formal library and participate in large-scale, sustained formalization efforts. Our full Lean code is currently being simplified by human experts. We just present the formal statement of the main theorem in Appendix~\ref{fs_of_BL}.

When formalising more involved arguments, the agent sometimes chose to further decompose the proof, introducing additional intermediate lemmas that were more fine-grained than those in the original blueprint. This behaviour appears to be a form of adaptive proof decomposition tailored to the demands of formal verification.

Moreover, compared with other specialized prover models, our agent is not restricted to theorem proving alone, but instead exhibits strong general-purpose reasoning capabilities. For a given formal statement whose correctness is not known in advance, traditional approaches typically can only attempt to prove both the original statement and its negation in parallel. In contrast, during our formalization of the Brascamp–Lieb inequalities, we observed that our agent is able to actively reason about the validity of the statement itself during the proof process. When it detects that a statement is incorrect, it can autonomously revise the statement accordingly. This ability to dynamically inspect and modify the problem formulation during formalization has not been present in previous provers. We present concrete examples of this behavior in the Appendix~\ref{self-correction}.

\subsection{Limitation}



Sometimes, our system generates Lean code that is overly long or less well-structured. In practice, we mainly tasked the agent with two kinds of 'sorry's of different difficulty. When a 'sorry' involved only local reasoning within an already well-structured proof, the agent usually filled the gaps with high-quality code. However, when a 'sorry' corresponded to the complete proof of a lemma, the agent was generally able to achieve the goal, but the resulting code tended to be verbose and less concise than desired. This highlights a limitation of the system: while it can handle complex proof goals, the readability and structure of generated formalizations may degrade for larger or more intricate tasks.


Our system occasionally struggles with type-level issues, which can significantly slow down the proof process. For instance, in one case, the agent failed completely—not because of difficulties in the core mathematical argument, but due to a type conversion from Real to NNReal. Such type-level constraints are rarely made explicit in informal mathematics, so the agent had difficulty reconstructing the required structure on its own. After revising the proof workflow and handling type conversions in advance to make the formalization path more “type-friendly”, the agent was able to complete the remaining proof successfully. This case highlights an inherent gap between informal and formal proofs and underscores the challenge that type-level requirements can pose for automated reasoning systems.


Moreover, despite their strong problem-solving capabilities in automated theorem proving, current agents still exhibit a clear gap between functional correctness and formal elegance. While agent-generated proofs often pass Lean’s compiler checks, they are frequently perceived by experienced Mathlib contributors as overly result-oriented, relying on verbose and low-level tactic scripts. Compared to human-written Mathlib code, these proofs lack structured abstraction and idiomatic use of higher-level patterns, leaving substantial room for improvement in conciseness, readability, and conformity to Mathlib’s community standards.

\clearpage

\bibliography{example_paper}
\bibliographystyle{icml2026}

\newpage
\appendix
\onecolumn
\section{Appendix}

\subsection{Formal statement of the main theorem of the Effective Brascamp–Lieb inequalities.}\label{fs_of_BL}

\begin{lstlisting}[frame = single]
theorem upperBound {J : Type*} [Fintype J]
    {E : Type*} [NormedAddCommGroup E] [InnerProductSpace ℝ E] [FiniteDimensional ℝ E]
    {F : J → Type*} [(j : J) → NormedAddCommGroup (F j)]
    [(j : J) → InnerProductSpace ℝ (F j)] [(j : J) → FiniteDimensional ℝ (F j)]
    (hE : Module.finrank ℝ E ≠ 0)
    (D : locRegDatum E F) (α : J → NNReal) (β : NNReal) (hα : ∀ i, 0 < α i)
    (hP : D.IsMetricPercep α β)
    (hS : ∀ j : J, (D.map j).EssentialRank rfl (α j) = Module.finrank ℝ (F j)) :
    let M_max := (D.loc + ∑ j, (D.weight j) • (D.map j).adjoint ∘ₗ (D.reg j) ∘ₗ (D.map j))
    loc_reg_constant_g D ≤
      (Module.finrank ℝ E : NNReal)^(D.Acuity / 2 : ℝ) *
      (Π j, (D.weight j)^(- (D.weight j : ℝ) * Module.finrank ℝ (F j) / 2)) *
      (Π j, (α j)^(- (D.weight j : ℝ) * Module.finrank ℝ (F j))) *
      ‖M_max.toContinuousLinearMap‖₊^((D.Acuity.toReal - Module.finrank ℝ E + β) / 2) *
      ‖(D.loc.equivOfDetNeZero D.pos_loc.2).symm.toContinuousLinearMap‖₊^(β.toReal / 2) := by sorry
\end{lstlisting}

\subsection{Self-Correction of Formal Statements during Formalization.}\label{self-correction}


\begin{lstlisting}[frame = single]
/-- When n is empty (dimension 0), the upper bound holds trivially.
    This lemma handles the degenerate case where the base space has dimension 0.
    In this case, both the LHS and RHS simplify to specific values and the inequality holds.
-/
lemma upperBound_empty_case {J : Type*} [Fintype J]
    {n : Type*} [Fintype n] [DecidableEq n]
    {m : J → Type*} [(j : J) → Fintype (m j)]
    (α : J → NNReal) (β : NNReal) (hα : ∀ i, 0 < α i)
    (D : locRegDatum (EuclideanSpace ℝ n) (fun j ↦ EuclideanSpace ℝ (m j)))
    (hP : D.IsMetricPercep α β)
    (hS : ∀ j : J, (D.map j).EssentialRank (α j) = Fintype.card (m j))
    (hn : ¬Nonempty n)
    (hβ_empty : β = 0)  -- Added: when n is empty, β must be 0 for the inequality to hold
    (A : (j : J) → EuclideanSpace ℝ (m j) →ₗ[ℝ] EuclideanSpace ℝ (m j))
    (hA : ∀ j, (A j).IsPosDef ∧ A j ≤ D.reg j) :
    (loc_constant_g_of n m D.1 A (fun j => (hA j).1) : ENNReal) ≤
      ↑((NNReal.rpow (Fintype.card n) (D.Acuity / 2)) *
      Π j, NNReal.rpow (D.weight j) (- D.weight j * Fintype.card (m j) / 2) *
      Π j, NNReal.rpow (α j) (- D.weight j * Fintype.card (m j)) *
      NNReal.rpow ‖(D.loc + ∑ j, (D.weight j) • (D.map j).adjoint ∘ₗ (D.reg j) ∘ₗ
        (D.map j)).toContinuousLinearMap‖₊
        ((D.Acuity - Fintype.card n + β) / 2) *
      NNReal.rpow
        ‖(D.loc.equivOfIsUnitDet (by simp [D.pos_loc.2])).symm.toLinearMap.toContinuousLinearMap‖₊
        (β / 2)) := by
\end{lstlisting}


\end{document}